\documentclass[11pt,a4paper]{article}
\usepackage[nohyperref]{acl2019}
\usepackage{times}
\usepackage{latexsym}
\usepackage{amsmath}
\usepackage{graphicx}
\usepackage[hyphens]{url}

\usepackage{url}

\aclfinalcopy 

\title{MSnet: A BERT-based Network for Gendered Pronoun Resolution}

\author{Zili Wang \\
  CEIEC \\
  Chengdu, China \\
  \texttt{wzlnot@gmail.com} \\}

\date{}

\begin{document}
\maketitle
\begin{abstract}
The pre-trained BERT model achieves a remarkable state of the art across a wide range of tasks in natural language processing.
For solving the gender bias in gendered pronoun resolution task, I propose a novel neural network model based on the pre-trained BERT.
This model is a type of mention score classifier and uses an attention mechanism with no parameters to compute the contextual representation of entity span, and a vector to represent the triple-wise semantic similarity among the pronoun and the entities.
In stage 1 of the gendered pronoun resolution task, a variant of this model, trained in the fine-tuning approach, reduced the multi-class logarithmic loss  to 0.3033 in the 5-fold cross-validation of training set and 0.2795 in testing set.
Besides, this variant won the 2nd place with a score at 0.17289 in stage 2 of the task.

The code in this paper is available at: \url{https://github.com/ziliwang/MSnet-for-Gendered-Pronoun-Resolution}\\
\end{abstract}

\section{Introduction}
Coreference resolution is an essential field of natural language processing \cite{review2018} and has been widely used in many systems such as dialog system \cite{cr_in_sa, wessel2017ontology}, relation extraction \cite{wang2018clinical} and question answer \cite{vicedo2000importance}.
Up to now, various models for coreference resolution have been proposed, and they can be generally categorized as (1) mention-pair classifier model \cite{webster2016vip}, (2) entity-centric model \cite{clark2015entity}, (3) ranking model  \cite{lee2017, lee2018}. However, some of these models implicate gender bias \cite{koolen2017these, rudinger2018gender}.
To address this, \citet{2018gap} presented and released Gendered Ambiguous Pronouns (GAP) dataset.

Recent work indicated that the pre-trained language representation models benefit to the coreference resolution \cite{lee2018}. 
In the past years, the development of deep learning methods of language representation was swift, and the newer methods were shown to have significant effects on improving other natural language processing tasks\cite{Peters2018, Radford2018, devlin2018bert}. 
The latest one is Bidirectional Encoder Representations from Transformers (BERT) \cite{devlin2018bert}, which is the cornerstone of the state of the art models in many tasks.

In this paper, I present a novel neural network model based on the pre-trained BERT for the gendered pronoun resolution task. 
The model is a kind of mention score classifier, and it is named as Mention Score Network (MSNet in short) and trained on the public GAP dataset.
In particular, the model adopts an attention mechanism to compute the contextual representation of the entity span, and a vector to represent the triple-wise semantic similarity among the pronoun and the entities.
Since the MSnet can not be tuned in a general way, I employ a two-step strategy to achieve the tuning-fine, which tunes the MSnet with freezing BERT firstly and then tunes them together. 
Two variants of MSnet are submitted in the gendered pronoun resolution task, and their logarithmic loss of local 5-fold cross-validation of train dataset is 0.3033 and 0.3042 respectively.
Moreover, in stage 2 of the task, they acquired the score at 0.17289 and 0.18361 respectively, by averaging the predictions on the test dataset, and won the 2nd place in the task.\\

\section{Model}
\label{sec:model}

As the target of the Gendered Pronoun Resolution task is to label the pronoun with whether it refers to entity A,  entity B, or NEITHER.  I aim to learn the reference probability distribution $P(E_i |D )$  from the input document $D$:
$$P(E_i|D)=\frac{\exp(s(E_i|D))}{\sum_{j \in E} \exp(s(E_j|D))}$$
where $E_i$ is the candidate reference entity of pronoun, $E = \{\text{A}, \text{B},\text{NEITHER}\}$ and $s$ is the score function which is implemented by a neural network architecture, which is described in detail in the following subsection.

\subsection{The Mention Score Network}
\begin{figure*}
  \includegraphics[width=\linewidth]{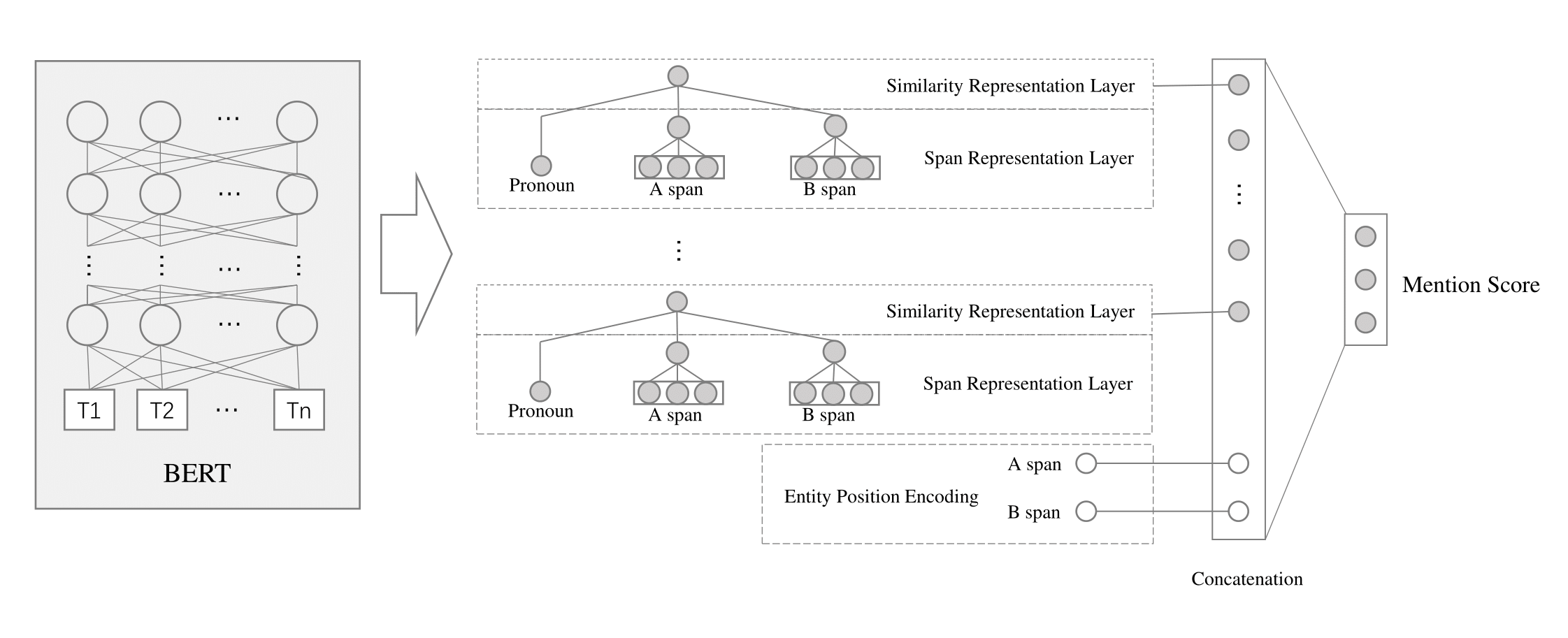}
  \caption{The architecture of MSnet.}
  \label{fig:msnet}
\end{figure*}

The mention score network is build on the pre-trained BERT model (Figure \ref{fig:msnet}).
It has three layers, the span representation layer, the similarity layer, and the mention score layer. They are described in detail in the following part.

\textbf{Span Representation Layer: }
The contextual representation is crucial to accurately predict the relation between the pronouns and the entities. 
Inspired by \citet{lee2017}, I adopt the hidden states of transformers of the pre-trained BERT as the contextual representation.
As \citet{devlin2018bert} showed that the performance of the concatenation of token representations from the top hidden layers of pre-trained Transformer of BERT is close to fine-tuning the entire model, the top hidden states will be given priority to compute the representation of entity spans.
Since most entity spans consist of various tokens, the contextual representation of them should be re-computed to maintain the correspondence.
I present two methods to re-compute the span representations:
1) \textbf{Meanpooling method}:
$$x^*_{(j, l)} = \frac{1}{\hat N}\sum_{i \in \text{Span}_j}x_{(i,l)}$$
where $x_{(i,l)}$ denotes the hidden states of $i$-th token in $l$-th layer of BERT, and $x^*_{(j , l)}$ denotes the contextual representation of entity span $j$, and $\hat N$ is the token counts of span $j$.
2) \textbf{Attention mechanism:}
Instead of weighting each token equality, I adopt the attention mechanism to weight the tokens by:
$$s_{(i, l)} = \frac{1}{\sqrt{d_H}}\text{norm}(x_{(i, l)}) \cdot x_{(p, l)}$$
$$a_{(i, j, l)} = \frac{\exp(s_{(i,l)})}{\sum_{k \in \text{Span}_j}\exp(s_{(k,l)})}\text{ , }i \in \text{Span}_j$$
$$x^*_{(j, l)}=\sum_{i \in \text{Span}_j}a_{(i, j, l)}x_{(i, l)}$$
The weights $a_{(i,j,l)}$ are learned automatically from the contextual similarity $s_{(i, l)}$ between pronoun $x_{(p, l)}$  and the token $x_{(i,l)}$ in the span $j$.
Different from the commonly used attention functions,  the above one has no parameters and is more space-efficient in practice.
The scaling factor $d_H$ denotes the hidden size of BERT and is designed to counteract the effect of extremely small gradients caused by the large magnitude of dot products \cite{Vaswani2017}.

\textbf{Similarity Layer: }
Inspired by the pairwise similarity of \citet{lee2017}, I assume a vector $\mathbf{\hat s}_l$ to represent the triple-wise semantic similarity among the pronoun and the entities of $l$-layer in BERT:
$$a_l = x^*_{(a, l)}$$
$$b_l= x^*_{(b, l)}$$
$$p_l=x_{(p,l)}$$
$$\mathbf{\hat s}_l=\mathbf{W}^T[p_l, a_l, b_l, a_l \circ p_l, b_l \circ p_l] + \mathbf{b}$$
where $a_l$, $b_l$ and $p_l$ denote the contextual representation of the pronoun, entity A  and entity B of the $l$-th layer in BERT, $\cdot$ denotes the dot product and $\circ$ denotes the element-wise multiplication.
The $\mathbf{\hat s}_l$ can be learned by a single layer feed-forward neural network with the weights $\mathbf{W}$ and the bias $\mathbf{b}$.

\textbf{Mention Score Layer: }
Mention score layer is also a feed-forward neural network architecture and computes the mention scores given the distance vector $\mathbf{d}$ between the pronoun and its candidate entities and the concatenated similarity vector $\mathbf{\hat s} $:
$$d_a = \tanh(w_{\text{dist}}(\text{START(A)$-$START(P)})+b_{\text{dist}})$$
$$d_b = \tanh(w_{\text{dist}}(\text{START(B)$-$START(P)})+b_{\text{dist}})$$
$$\mathbf{d} = [d_a, d_b]$$
$$ \mathbf{\hat s}=[\mathbf{\hat s}_0,\mathbf{\hat s}_1,..., \mathbf{\hat s}_l, ... ,\mathbf{\hat s}_L]$$
$$s(E_i|D) = \mathbf{W}_{E_i} \cdot [\mathbf{\hat s} , \mathbf{d} ]+b_{E_i}$$
where $d_a$ (or $d_b$) denotes the distance encoding of entity A (or B), $\mathbf{\hat s}_l$ denotes the similarity vector computed by the representation of the $l$-th layer in BERT.
$L$ is the total layers for representation, and $\text{START}$ denotes the index of the start token of the span.
$w_{\text{dist}}$ is a learnable weight for encoding the distance which corresponds to a learnable bias $b_{\text{dist}}$ and $\mathbf{W}_{E_i}$ is the learnable weights for scoring entity $E_i $ which corresponds to a learnable bias $b_{E_i}$.\\

\section{Experiments}
I train the model on the Kaggle platform by using scripts kernel which using the computational environment from the docker-python \footnote{\url{https://github.com/Kaggle/docker-python}}.
I employ pytorch as the deep learning framework, and the \verb|pytorch-pretrained-BERT| package \footnote{\url{https://github.com/huggingface/pytorch-pretrained-BERT}} to load and tune the pre-trained BERT model.

\subsection{Dataset}
The GAP Coreference Dataset \footnote{\url{https://github.com/google-research-datasets/gap-coreference}} \cite{2018gap} has 4454 records and officially split into three parts: development set (2000 records), test set (2000 records), and validation set (454 records).
Conforming to the stage 1 of Gendered Pronoun Resolution \footnote{\url{https://www.kaggle.com/c/gendered-pronoun-resolution}} task, the official test set and validation set are combined as the training dataset in the experiments,  while the official development set is used as the test set correspondingly.

\subsection{Preprocessing}
In the experiments, the WordPiece is used to tokenize the documents. To ensure the token counts less than 300 after tokenizing, I remove the head or tail tokens in a few documents. Next, the special tokens \verb|[CLS]| and \verb|[SEP]| are added into the head and end of the tokens sequences.

\subsection{Hyper-parameters}
\label{sub:hyper}
\textbf{Pre-trained BERT model}:
As increasing model sizes of BERT may lead to significant improvements on very small scale tasks \cite{devlin2018bert}, I explore the effect of BERT$_\text{BASE}$ and BERT$_\text{LARGE}$ in the experiments.
I employ the \verb|uncased_L-12_H-768_A-12| \footnote{\url{https://storage.googleapis.com/bert_models/2018_10_18/uncased_L-12_H-768_A-12.zip}} as the BERT$_\text{BASE}$ and \verb|cased_L-24_H-1024_A-16| \footnote{\url{https://storage.googleapis.com/bert_models/2018_10_18/cased_L-24_H-1024_A-16.zip}} as the BERT$_\text{LARGE}$, and both of them are transformed into the pytorch-supported format by the script in \verb|pytorch-pretrained-BERT|. 

\textbf{Hidden Layers for Representation}:
\citet{devlin2018bert} showed that using the representation from appropriate hidden layers of BERT can improve the model performance, the hidden layers $L$ (described in Section ~\ref{sec:model}) is  therefore utilized as a hyper-parameter tuned in the experiments.

\textbf{Dimension of Similarity Vector}: 
Since a vector is used to represent the task-specific semantic similarity, its dimension $\mathbf{\hat s}_{dim}$ may have potential influence the performance.
A smaller dimension will partly lose information, while a bigger one will cause generalization problems.

\textbf{Span Contextual Representation}:
As section \ref{sec:model} described, both the meanpooling and attention method can be used to compute the contextual representation of the tokens span of the entity.
Therefore, the choice of them is a hyper-parameter in the experiment.

\textbf{Tunable Layers}:
I use two different approaches to train the MSnet model. The first one is the feature-based approach which trains MSnet  with  freezing the BERT part. The second one is the fine-tuning approach, which tunes the parameters of BERT and MSnet simultaneously.  
\citet{howard2018universal} showed the discriminative fine-tuning gets a better performance than the ordinary, which possibly means that the pre-trained language model has a hierarchical structure.
One possible explanation is that the lower hidden layers extract the word meanings and grammatical structures and the higher layers process them into higher-level semantic information.
In this, I freeze the embedding layer and bottom hidden layers of BERT to keep the completeness of word meaning and grammatical structure and tune the top hidden layers $L_\text{tuning}$.

\subsection{Training Details}
For improving the generalization ability of the model, I employ the dropout mechanism \cite{srivastava2014dropout} on the input of the feed-forward neural network in the similarity layer and the concatenation in the mention score layer. The rate of dropout is set at 0.6 which is the best setting after tuned on it.
I also apply the dropout on the representation of tokens when using the attention mechanism to compute the contextual representation of span, and its dropout rate is set at 0.4.
Additionally, I adopt the batch normalization \cite{2015batch} before the dropout operation in the mention score layer.
As introduced in section \ref{sub:hyper}, I use the feature-based approach and the fine-tuning approach separately to train the MSnet, and the training details are described in the following.
 
\textbf{Feature-based Approach}: 
In the feature-based approach, I train the model by minimizing the cross-entropy loss with Adam \cite{2014adam} optimizer with a batch size of 32. To adapt to the training data in the experiments, I tuned the learning rate and found a learning rate of 3e-4 was the best setting. The maximum epoch set at 30 and early stopping method is used to prevent the over-fitting of MSnet.

\textbf{Fine-tuning Approach}:
In the fine-tuning approach, the generic training method was not working.
I adopt a two-step tuning strategy to achieve the fine-tuning.
In step 1, I train the MSnet in the feature-based approach.
And in step 2,  MSnet and BERT are tuned simultaneously with a small learning rate.

Since the two steps have the same optimization landscape, in step 2, the model may not escape the local minimum where it entered in step 1.
I adopt two strategies of training in step 1 to reduce the probability of those situations: 1) premature. The MSnet is trained to under-fitting by using a small maximum training epoch which is set at 10 in the experiments. 2) mature. In this strategy, MSnet is trained to proper-fitting, and it is applied by adopting a weight decay at 0.01 rate, an early stopping at 4 epoch, and the maximum training epoch at 20 in the experiments.
In addition, other training parameters of the two strategies have the same setting as in the feature-based approach.

In step 2, I also trained the model by minimizing the cross-entropy loss but with two different optimizers.
For BERT, I used the Adam optimizer with the weight decay fix which implemented by \verb|pytorch-pretrained-BERT|.
For MSnet, the generic Adam was used. Both of the two optimizers are set with a learning rate at 5e-6 and a weight decay at 0.01.
The maximum training epoch is set at 20, and the early stopping is set at 4 epoch.
The batch size was 5 as the GPU memory limitation.

\subsection{Evaluation}
I report the multi-class logarithmic loss of the 5-fold cross-validation on train and the average of their predictions on the test.  Also, the running time of the scripts is reported as a reference of the performance of the MSnet. \\

\section{Results and Discussion}

\begin{table*}[t!]
\centering
\begin{tabular}{c|c|c|c|c|c|c|c}
  \hline
  Model\verb|#| & BERT & $L$ & $\mathbf{\hat s}_\text{dim}$ & Span & 5-fold CV on train& test & runtime(s) \\
  \hline
  1 & BASE & 1 & 32 & Meanpooling & 0.5247$\pm$0.0379 & 0.4891 & 232.8 \\
  2 & BASE & 4 & 32 & Meanpooling & 0.4699$\pm$0.0431 & 0.4270 &317.3\\
 \hline
  3 & LARGE &4 & 32 & Meanpooling & 0.4041$\pm$0.0532 &0.3819 &358.3\\
  4 & LARGE &8 &32 &Meanpooling &0.3783$\pm$0.0468 &0.3519 &372.2\\
  5 & LARGE &12 &32 &Meanpooling &0.3879$\pm$0.0461& 0.3546 &415.4\\
\hline
  6 & LARGE &8 &8 &Meanpooling &0.3758$\pm$0.0430&0.3490&436.2\\
  7 & LARGE &8 &16 &Meanpooling &0.3736$\pm$0.0465&0.3488&415.0\\
  8 & LARGE &8 & 64&Meanpooling &0.3780$\pm$0.0441&0.3518&447.6 \\
\hline
  9 & LARGE & 8 & 16 & Attention & 0.3582$\pm$0.0435&0.3349&828.2\\
\hline
\end{tabular}
\caption{Results of Feature-based Aproach.}
\label{tab:feat}
\end{table*}

\begin{table*}[t!]
\centering
\begin{tabular}{c|c|c|c|c|c|c}
\hline
Model\verb|#| & Based Model & method & $L_{\text{tuning}}$ & 5-fold CV on train & test & runtime(s) \\
\hline
 10 & \verb|#|9 & premature & 12& 0.3033$\pm$0.0367 &0.2795&6909.5\\
 11 & \verb|#|9	& mature & 12 &	0.3042$\pm$0.0352 & 0.2856 & 7627.7\\
\hline
12 & \verb|#|9 & mature & 8	& 0.3110$\pm$0.0352 & 0.2876 & 8928.1 \\
13 & \verb|#|9 & mature & 16 & 0.3185$\pm$0.0465 & 0.2820 & 7763.4 \\
14 & \verb|#|9 & mature & 24 & 0.3169$\pm$0.0440 & 0.2843 & 8695.4 \\
\hline
\end{tabular}
\caption{Results of Fine-tuning Aproach.}
\label{tab:tuning}
\end{table*}

\subsection{Feature-based Approach}
The results of MSnet variants trained in feature-based approach are shown in Table \ref{tab:feat}.
The comparison between model  \verb|#|1 and model \verb|#|2 shows that the combination of the top 4 hidden layers for contextual representation is better than the top layer.
The possible reason is that the semantic information about gender may be partly transformed to the higher level semantic information during the hidden layers in BERT.
In addition, changing BERT$_\text{BASE}$ to the BERT$_\text{LARGE}$ reduces the loss in 5-fold CV on train from 0.4699$\pm$0.0431 to 0.4041$\pm$0.0532, which demonstrate increasing model size of BERT can lead to remarkable improvement on the small scale task.
The exploration of contextual representation layers shows the proper representation layers is proportionate to the number of hidden layers of BERT.
In other words, the modeling ability of  BERT$_\text{LARGE}$ is more powerful than BERT$_\text{BASE}$ by using a more complex function to do the same work.

The comparison among the model \verb|#|4, model \verb|#|6, model \verb|#|7 and model \verb|#|8 shows the dimension of the similarity vector has a slight affection for the performance of MSnet (Table \ref{tab:feat}) and the best loss is 0.3736$\pm$0.0465 with the dimension set at 16.
Changing the method for computing the span contextual representation from meanpooling to attention mechanism reduces the loss in CV on train by $\sim$0.02, which demonstrates that the attention mechanism used in the experiment is effective to compute the contextual representation of the entity span. To the best of my knowledge, it is a novel attention mechanism with no learnable parameters and more space-efficient and more explainable in practice.

\subsection{Fine-tuning Approach}
The experiments in fine-tuning approach was based on model \verb|#|9, and the results are shown in table \ref{tab:tuning}.
The comparison between model \verb|#|10 and model \verb|#|11 shows that their difference on performance is slight.
Also,  both of them are effective to the fine-tuning of MSnet and reduce loss in the CV of train by $\sim$0.054 compared to the feature-based approach. 
Furthermore, the tuning on $L_{\text{tuning}}$ shows the best setting is tuning top 12 hidden layers in BERT, and more or fewer layers will reduce the performance of MSnet.
The possible reason is that tuning fewer layers will limit the ability of the transformation from basic semantic to gender-related semantic while tuning more bottom layers will damage the extraction of the underlying semantics when training on a small data set.

As the apporach transformed from the feature-based to the fine-tuning,  the intentions of some hyper-parameters were changed.
The obvious one is the hidden layers for contextual representation, which is used to combine the semantic in each hidden layers in the feature-based approach and changed to constrain the contextual representation to include the same semantic in fine-tuning approach.
Although, the change on the intentions was not deliberate, the improvement on the performance of the model was observed in the experiments.

\subsection{Results in Stage 2}
The gendered pronoun resolution was a two-stage task, and I submitted the model \verb|#|10 and \verb|#|11 in stage 2 as their best performances in 5-fold cross-validation of the training dataset.
The final scores of the models were 0.17289 (model \verb|#|10 ) and 0.18361 (model  \verb|#|11).
This result featurely demonstrates the premature strategy is better than the mature one and can be explained as former one keeps more explorable optimization landscape in step 2 in the fine-tuning approach.\\

\section{Conclusion}
This paper presented a novel pre-trained BERT based network model for the gendered pronoun resolution task.
This model is a kind of mention score classifier and uses an attention mechanism to compucate the contextual representation of entity span and a vector to represent the triple-wise semantic similarity among the pronoun and the entities.
I trained the model in the feature-based and the two-step fine-tuning approach respectively.
On the GAP dateset, the model trained by the fine-tuning approach with premature strategy obtains remarkable multi-class logarithmic loss on the local 5-fold cross-valication at 0.3033, 
and 0.17289 on the test dataset in stage 2 of the task.
I believe the MSnet can serve as a new strong baseline for gendered pronoun resolution task as well as the coreference resolution.
The code for training model are available at: \url{https://github.com/ziliwang/MSnet-for-Gendered-Pronoun-Resolution} \\

\section*{Acknowledgments}
I want to thank the kaggle company for its public computing resources. \\

\bibliography{myacl2019}
\bibliographystyle{acl_natbib}

\end{document}